\documentclass[conference]{IEEEtran}
\pdfoutput=1
\IEEEoverridecommandlockouts

\usepackage{graphicx}
\usepackage{color}

\usepackage{cite}
\usepackage{amsmath}
\usepackage[caption=false]{subfig}

\usepackage{amssymb}

\usepackage{comment}

\usepackage[acronym]{glossaries}

\newcommand{\erm}{\textbf{ERM}}
\newcommand{\irm}{\textbf{IRM}}
\newcommand{\best}{\textbf{BEST}}
\newcommand{\random}{\textbf{RAND}}

\newcommand{\env}[1]{ENV\#{#1}}
\newcommand{\fractionEnv}{\alpha_E}
\newcommand{\fractionCls}{\alpha_C}

\newcommand{\lossFn}{\ell}
\newcommand{\lossCoef}{\ell_0}
\newcommand{\representFn}{T}
\newcommand{\regularizeCoef}{\lambda}

\newcommand{\element}{k}
\newcommand{\ELEMENT}{K}
\newcommand{\elementSet}{\set{\ELEMENT}}

\newcommand{\action}{\psi}
\newcommand{\actionVec}{\vect{\action}}

\newcommand{\actionVecEst}{\estimated{\actionVec}}
\newcommand{\actionVecOpt}{\vect{\actionOpt}}

\newcommand{\actionEst}{\estimated{\action}}
\newcommand{\actionOpt}{\phi}

\newcommand{\signalTx}{v}
\newcommand{\signalRx}{y}
\newcommand{\noiseTx}{\tilde{n}}

\newcommand{\distance}{d}

\newcommand{\response}{\vect{a}}
\newcommand{\radiation}{G}
\newcommand{\pathloss}{L}

\newcommand{\distanceTx}{\distance_{\text{RIS},\text{T}}}
\newcommand{\distanceRx}{\distance_{\text{RIS},\text{R}}}

\newcommand{\azimuth}{\phi}
\newcommand{\azimuthTx}{\azimuth_{\text{RIS},\text{T}}}
\newcommand{\azimuthRx}{\azimuth_{\text{RIS},\text{R}}}

\newcommand{\elevation}{\theta}
\newcommand{\elevationTx}{\elevation_{\text{RIS},\text{T}}}
\newcommand{\elevationRx}{\elevation_{\text{RIS},\text{R}}}

\newcommand{\scatter}{s}
\newcommand{\SCATTER}{S}
\newcommand{\scatterSet}{\set{\SCATTER}}

\newcommand{\configurationSet}{\set{C}}

\newcommand{\rate}{r}

\newcommand{\channelIn}{h}
\newcommand{\channelInVec}{\vect{\channelIn}}
\newcommand{\addLos}[1]{#1_{\text{LOS}}}
\newcommand{\addNlos}[1]{#1_{\text{NLOS}}}

\newcommand{\channelOut}{g}
\newcommand{\channelOutVec}{\vect{\channelOut}}

\newcommand{\mapping}{f}

\newcommand{\environment}{e}
\newcommand{\ENVIRONMENT}{E}
\newcommand{\environmentSet}{\set{\ENVIRONMENT}}

\newcommand{\inputOrg}{x}
\newcommand{\inputOrgVec}{\vect{\inputOrg}}

\newcommand{\inputMod}{z}
\newcommand{\inputModVec}{\vect{\inputMod}}

\newcommand{\noiseAlone}{ N_0}
\newcommand{\noise}{ \noiseAlone}
\newcommand{\bandwidth}{W}

\newcommand{\txpower}{p}

\newcommand{\dataSet}{\set{D}}
\newcommand{\sample}{n}
\newcommand{\SAMPLE}{N}

\newcommand{\estimated}[1]{\hat{#1}}

\newcommand{\modelParam}{\vect{w}}
\newcommand{\modelParamInd}{w}

\newcommand*{\myfigfactor}{0.8}
\newcommand*{\myfigfactorx}{0.31}

\DeclareMathOperator*{\argmin}{arg\,min}

\newcommand{\optMinimize}{\min} 
\newcommand{\optMaximize}{\max} 
\newcommand{\subjectTo}{\text{s. t.}}

\newcommand{\expect}{\mathbb{E}\,}
\newcommand{\probability}{\text{Pr}}

\newcommand{\normalDistribution}{\mathcal{N}}
\newcommand{\normalDistributionComplex}{\mathcal{CN}}

\newcommand{\uniformDistribution}{\mathcal{U}}

\DeclareMathOperator*{\abs}{abs}

\newcommand{\realset}{\mathbb{R}}

\newcommand{\vect}{\boldsymbol}

\newcommand{\seta}[1]{1,\dots,#1}
\newcommand{\set}[1]{\mathcal{#1}}

\newcommand{\one}{\mathbf{1}}
\newcommand{\zero}{\mathbf{0}}

\newcommand{\indictsimp}[1]{\mathbb{I}({#1})}

\newcommand{\transpose}{^\dag}

\newcommand{\grad}[2]{\nabla_{#1}#2}
\newcommand{\daba}[2]{\frac{\partial #1}{\partial #2}}

\newcommand{\sciNotation}[2]{#1\times 10^{#2}}

\usepackage{forloop}
\newcounter{loopcntr}
\newcommand{\rpt}[2][1]{%
	\forloop{loopcntr}{0}{\value{loopcntr}<#1}{#2}%
}

\newcommand{\subgroup}[1]%
{\rlap{\smash{%
	\newcount\cnt%
	\cnt \numexpr#1\relax%
	\advance\cnt -1\relax%
	$\tabcolsep=.1em\begin{tabular}[t]{|l}\multicolumn{1}{l}{}\\%
	\rpt[\cnt]{\\}
	\\\hline\end{tabular}$%
}}}

\newcounter{myRefCount}

\newacronym{5g}{5G}{fifth generation}

\newacronym{tx}{Tx}{transmitter}

\newacronym{rx}{Rx}{receiver}

\newacronym{ris}{RIS}{reconfigurable intelligent surface}

\newacronym{csi}{CSI}{channel state information}

\newacronym{mle}{MLE}{maximum likelihood estimation}

\newacronym{ml}{ML}{machine learning}

\newacronym{mmw}{mmWave}{millimeter wave}

\newacronym{los}{LOS}{line-of-sight}

\newacronym{nlos}{NLOS}{non line-of-sight}

\newacronym{rss}{RSS}{received signal strength}

\newacronym{snr}{SNR}{signal to noise ratio}

\newacronym{pdf}{PDF}{probability distribution function}

\newacronym{cdf}{CDF}{cummulative density function}

\newacronym{gpr}{GPR}{Gaussian process regression}

\newacronym{nn}{NN}{neural network}

\newacronym{cnn}{CNN}{convolutional neural network}

\newacronym{relu}{ReLU}{rectified linear unit}

\newacronym{erm}{ERM}{empirical risk minimization}

\newacronym{irm}{IRM}{invariant risk minimization}

\newacronym{aoa}{AoA}{angle of arrival}

\newacronym{aod}{AoD}{angle of departure}

\newacronym{mlp}{MLP}{multi layer perceptrons}

\newacronym{ood}{OOD}{out-of distribution}
\allowdisplaybreaks

\begin{document}

\pagenumbering{gobble}
\title{%
	Robust Reconfigurable Intelligent Surfaces via Invariant Risk and Causal Representations 
}

\author{
\IEEEauthorblockN{
	Sumudu Samarakoon\IEEEauthorrefmark{1},
	Jihong Park\IEEEauthorrefmark{2}, 
	and
	Mehdi Bennis\IEEEauthorrefmark{1}
	\\}
\IEEEauthorblockA{
	\small%
	\IEEEauthorrefmark{1}%
	Centre for Wireless Communication, University of Oulu, Finland, email: \{sumudu.samarakoon,mehdi.bennis\}@oulu.fi \\
	\IEEEauthorrefmark{2}%
	School of Information Technology, Deakin University, Geelong, VIC 3220, Australia, email: jihong.park@deakin.edu.au
}
}

\maketitle
\nopagebreak[4]

\begin{abstract}

In this paper, the problem of robust \gls{ris} system design under changes in data distributions is investigated.
Using the notion of \gls{irm}, an invariant causal representation across multiple environments is used such that the predictor is simultaneously optimal for each environment.
A neural network-based solution is adopted to seek the predictor and its performance is validated via simulations against an empirical risk minimization-based design. 
Results show that leveraging invariance yields more robustness against unseen and out-of-distribution testing environments.
\end{abstract}

\glsresetall
\section{Introduction}\label{sec:introduction}

\Glspl{ris}  have   recently  gained  remarkable  attention  as  a  low-cost,  hardware-efficient, and highly scalable technology capable of offering dynamic control of electromagnetic wave propagation \cite{He2020,Oezdogan2020,Chen2019}.
Due to the design based on nearly passive multiple reflective elements, the channel acquisition and the dynamic configuration of \gls{ris} parameters are two main challenges in \gls{ris}-assisted wireless communication.

The overwhelming majority of the existing literature on communication aided by \glspl{ris} relies on the availability of perfect/estimated \gls{csi} to develop \gls{ml}-based \gls{ris} configuration \cite{Oezdogan2020,Chen2019,Gao2020,Lee2020}.
Therein, frequent channel sampling via dedicated control signaling and pilot-based estimation techniques that introduce additional overhead and communication complexity to the system under channel dynamics have been used.
To reduce such overheads imposed by the need of perfect \gls{csi}, 
the works of \cite{Park2020,Sheen2021,Alexandropoulos2020} have developed \gls{nn}-based \gls{ris} configuration solutions utilizing system properties such as locations and relative distances of transmitters, receivers, and \glspl{ris} instead of relying on \gls{csi} measurements.
All current designs mainly focus on exploiting statistical correlations within the observed data.
However, these works neglect the data generation process and the underlying causal relationships between the environment and \gls{ris} configurations.
Moreover, current approaches yield high inference accuracy for a given environment, mostly when the training and test data distribution are identical.
As a result, these approaches fail to generalize \gls{ood} data and other unseen environmental changes.

The main contribution of this article is to fill this void and develop \textbf{a novel robust learning framework for predicting \gls{ris} configurations} rooted in learning over representations that are invariant across different environments.
Towards this goal, it is essential to discover the causal dependencies of system components (e.g., \gls{ris} configuration, angles of arrivals/departures at transmitters, receivers, and reflectors) that are invariant over different environments, i.e., \emph{invariant representation of network parameters} in contrast to exploiting environment-dependent spurious correlations (e.g., scatterer distribution affecting \gls{csi}). 
In this work, we use a supervised learning approach to determine the optimal \gls{ris} configurations based on \gls{csi} and network-wide parameters pairs collected over different network settings (referred to as \emph{environments}).
The conventional approach is to seek predictors based on the observed \gls{csi} that minimizes the empirical loss; a process known as \gls{erm}.
In contrast, we formulate an \gls{irm} problem by casting the prediction of \gls{ris} phase configuration as a loss minimization problem leveraging an invariant representation of the data (\gls{csi}) that is optimal for all environments.
Based on the \gls{irm} formulation, we train a \gls{nn}-based robust \gls{ris} configuration predictor and show that the proposed predictor outperforms an \gls{erm}-based predictor in terms of \gls{ood} generalization and unseen test environments with up to $15\%$ higher prediction accuracy.

The rest of the paper is organized as follows.
Section \ref{sec:system_model} describes the system model and the conventional design of the optimal \gls{ris} phase configuration predictor using \gls{erm}.
The \gls{irm} based \gls{ris} configuration design that utilizes learning over invariant representation is discussed in  \ref{sec:irm}.
Section \ref{sec:results} evaluates and compares the proposed solution with the baselines by means of extensive simulations in terms of generalization and \gls{ood} robustness.
Finally, conclusions are drawn in Section \ref{sec:conclusion}.

\begin{figure*}
	\centering
	\includegraphics[width=.95\linewidth]{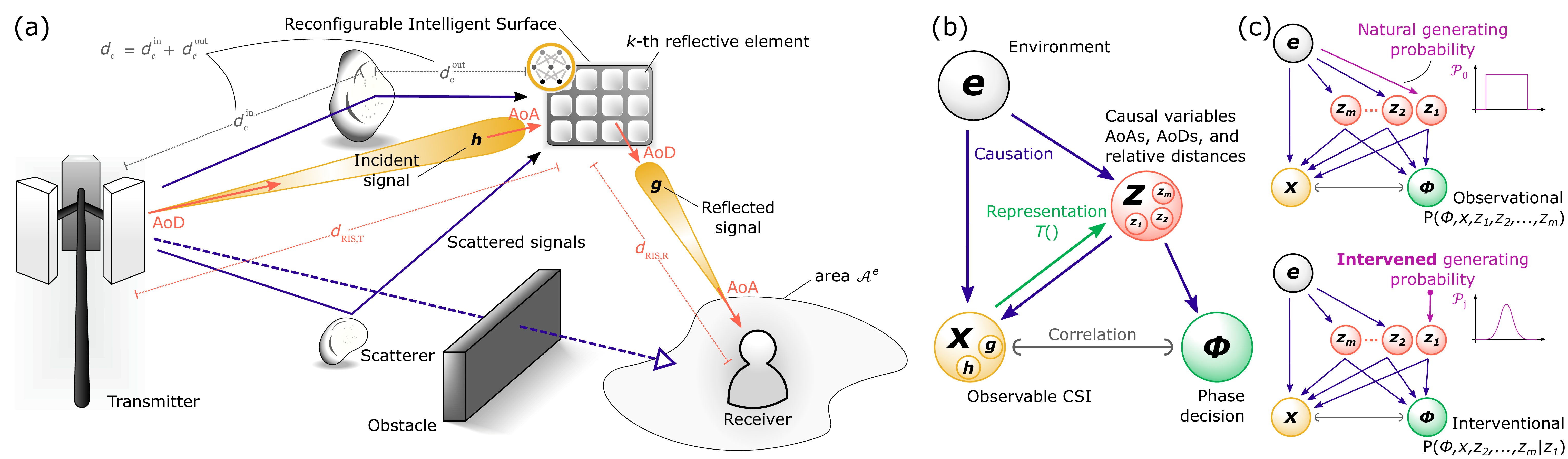}
	\caption{System model illustrating \glspl{ris}-aided \gls{tx}-\gls{rx} connectivity under blocked line-of-sight within an instance of environment $\environment$ is given in (a). The causal relation among the observable features, decision, and the environment along the observable correlation is illustrated under (b). The impact of intervening the generating probability of $\inputModVec_1$ yielding an intervention outcome rather than of the observable one is illustrated in (c). }
	\label{fig:combined}
\end{figure*}

\section{System Model \& Problem Formulation}\label{sec:system_model}

As illustrated in Fig.~\ref{fig:combined}, we consider a set of multiple downlink \gls{ris} communication environments $\environmentSet$ defined from the \gls{rx}'s perspective, such that an environment $\environment\in\environmentSet$ consists of a single \gls{tx}-\gls{rx} pair without \gls{los}, an \gls{ris}, randomly located scatterers in the \gls{tx}'s vicinity, and an area $\mathcal{A}^\environment$ that the \gls{rx} occupies. 
The \gls{ris} enables \gls{nlos} communication from \gls{tx} to \gls{rx}, via reflecting signals transmitted from \gls{tx} and diffracted signals from a set of scatterers ($\scatterSet$). 
Therein, the goal is to maximize the downlink data rate by adjusting the phases of the \gls{ris} elements. 
The system model and problem formulation are elaborated next.

\subsection{Channel model}

Let $\channelInVec = [\channelIn_{\element}]_{\element\in\elementSet}$ and $\channelOutVec = [\channelOut_{\element}]_{\element\in\elementSet}$ be the channel vectors of the incident (\gls{tx}-\gls{ris}) and reflected (\gls{ris}-\gls{rx}) signals defined over the reflective elements $\elementSet$ in the \gls{ris}.
The channel model is based on the work in \cite{Basar2020}, in which, the link between \gls{tx} and \gls{ris} is composed of \gls{los} channels as well as \gls{nlos} channels due to the presence of scatterers, while the \gls{ris}-\gls{rx} link has \gls{los} connectivity due to their close proximity.
Let $\distance_o$, $\azimuth_o$, and $\elevation_o$ be the distance, azimuth angle, and elevation angle of an object $o\in\{\text{\gls{tx}},\text{\gls{rx}}\}\cup\scatterSet$ with respect to the \gls{ris}.
With a uniformly distributed random phase $\eta\sim\uniformDistribution[0,2\pi]$ and $\imath^2=-1$, under the assumption that the scatterers are only in the vicinity of the \gls{tx}, the \gls{ris}-\gls{rx} channel is modeled as follows:
\begin{equation}\label{eqn:channel_ris_rx}
	\channelOutVec = \sqrt{\radiation(\elevationRx) \pathloss(\distanceRx)}
	e^{\imath\eta} \response(\azimuthRx,\elevationRx),
\end{equation}
where $\radiation(\cdot)$, $\pathloss(\cdot)$, and $\response(\cdot)$ are the \gls{ris} element radiation pattern, distance-dependent path loss, and array response, respectively \cite{Basar2020}.
Similar to \eqref{eqn:channel_ris_rx}, the \gls{los} component of the channel between \gls{ris} and \gls{tx} is modeled by,
\begin{equation}\label{eqn:channel_ris_tx_los}
	\addLos{\channelInVec} = 
	\sqrt{\radiation(\elevationTx) \pathloss(\distanceTx)}
	e^{\imath\eta} \response(\azimuthTx,\elevationTx),
\end{equation}
as defined in the \gls{5g} channel model \cite{Basar2020}.

The \gls{nlos} links between the \gls{tx} and \gls{ris} are due to the presence of scatterers.
Let $\distance_{\scatter}$, $\azimuth_\scatter$, and $\elevation_\scatter$ be the  traveled-distance of the reflected signal from \gls{tx} to \gls{ris} at scatterer $\scatter$ and the azimuth and elevation angles of scatterer $\scatter$ with respect to the \gls{ris}, respectively.
Then, the \gls{nlos} channel is modeled as follows;
\begin{equation}\label{eqn:channel_ris_tx_nlos}
	\addNlos{\channelInVec} = 
	\frac{1}{\SCATTER} \sum_{\scatter} \vect{\gamma}_\scatter
	\sqrt{ \radiation(\elevation_\scatter) \pathloss(\distance_\scatter)}
	\response(\azimuth_\scatter,\elevation_\scatter),
\end{equation}
where $\gamma_\scatter\sim\normalDistributionComplex(\zero,\one)$ is a scatterer-dependent random path gain.
In this view, the channel between \gls{tx} and \gls{ris} becomes $\channelInVec = \addLos{\channelInVec} + \addNlos{\channelInVec}$.

\subsection{Rate maximization}

At the \gls{ris}, the phases of incident signals are altered to enhance the capacity at the \gls{rx}.
Denoting the phase change decision of the \gls{ris} over its reflective elements by $\actionVec = [\action_{\element}]_{\element\in\elementSet}$ with $\abs( \action_{\element} ) = 1$, the received signal $\signalRx$ at the \gls{rx} is given by,
\begin{equation}\label{eqn:signal_received}
	\signalRx = 
	\textstyle  \channelOutVec\transpose \actionVec {\channelInVec} \signalTx + \noiseTx,
\end{equation}
where $\signalTx$ is the transmit signal with $\expect[\signalTx^2]=\txpower$
and
$\noiseTx\sim\normalDistribution(0,\noiseAlone)$ is the noise.
The data rate at the \gls{rx} is 
$\rate(\actionVec,\channelInVec,\channelOutVec) = \bandwidth \log_2 \big( 1 + 
\textstyle \frac{| \channelOutVec\transpose \actionVec {\channelInVec} |^2 \txpower }{\bandwidth\noise} \big)$
where $\bandwidth$ is the bandwidth
and
In this view, the data rate maximization at the \gls{rx} is cast as follows:
\begin{eqnarray}
	\label{eqn:maximize_datarate}
	\underset{\actionVec\in\configurationSet}{\optMaximize} && 
	\rate(\actionVec,\channelInVec,\channelOutVec) = 
	\bandwidth \log_2 \big( 1 + 
	\textstyle \frac{|  \channelOutVec\transpose \actionVec {\channelInVec} |^2 \txpower }{\bandwidth\noise} \big),
\end{eqnarray}
where $\configurationSet$ is the feasible set of \gls{ris}  configurations.

Under the perfect knowledge of \gls{csi} over all links, the optimal phase decision
$\actionVecOpt$ can be obtained through an exhaustive search.
However, this poses two challenges: 
(i) the complexity of a heuristic search increases with the number of reflective elements and their configurations
and
(ii) assuming perfect \gls{csi} at \gls{ris} requires a huge number of channel measurements.
The issue (i) can be addressed by the aid of \gls{ml}-based regression.
Therein, given a dataset $\dataSet = \{ (\inputOrgVec,\actionVecOpt)_\sample | \sample\in\{\seta{\SAMPLE}\} \}$\footnote{
Note that the subscript $\sample$ is neglected from all parameters for simplicity unless the notion of sample is significant. 
}
consisting of observed \gls{csi} $\inputOrgVec=(\channelInVec,\channelOutVec)$ and optimal configuration labels, a mapping function $\tilde{\mapping}_{\modelParam}(\cdot)$ parameterized by $\modelParam$, referred to as a \emph{predictor}, $\actionVecEst = \tilde{\mapping}_{\modelParam}(\inputOrgVec)$ is obtained by solving the \gls{erm} as follows,
\begin{eqnarray}
	\label{eqn:erm_problem}
	\underset{\modelParam}{\optMinimize} && 
	\textstyle
	\frac{1}{\SAMPLE} \sum_{\sample} \lossFn(\actionVecOpt_\sample,\actionVecEst_\sample),
\end{eqnarray}
where $\lossFn(\actionVecOpt,\actionVecEst) = \sum_{\element} \lossCoef
\sin^2 \big( \frac{\actionEst_\element - \actionOpt_\element}{2} )$ is the loss (or risk) function in terms of phase prediction with an arbitrary scaling coefficient $\lossCoef>0$.
Nevertheless, due to the issue in (ii), training over a larger dataset collected over different channel realizations and network configurations, as well as embedding \gls{csi} measurement capabilities at \glspl{ris} for inference are impractical. 
In short the application of \erm{} for \glspl{ris} poses algorithmic and practical difficulties  in terms of lack of robustness and generalization across  different environments.
This calls for developing a new learning framework for predicting  phase configurations that are robust across multiple environments.

\section{\gls{irm}-based Phase Optimization}\label{sec:irm}

The main limitation of analyzing a limited set of observations is due to the resultant predictor $\tilde{\mapping}_{\modelParam}$ that depends on spurious correlations among observables. 
Overfitting to spurious correlations based on the knowledge of the scattered signals while neglecting
the underlying causal relations that are invariant across multiple environments prevents an \erm{}-based system design to operate as a robust predictor against changes in the environment.
Therefore, it is essential to develop an invariant predictor based on causality rather than exploiting spurious correlations.

Towards the design of an invariant predictor across multiple environments, we first identify $(\inputOrgVec,\actionVecOpt)_\sample^\environment$ as the $\sample$-th input-output tuple (sample) of environment $\environment$.
Hence, the entire dataset can be seen as a composite of data collected over different environments, i.e., $\dataSet = \cup_\environment \dataSet_\environment$.
Concretely, we seek a parameterized mapping function (predictor) $\actionVecEst_\sample^\environment = \mapping_{\modelParam}(\inputModVec_\sample^\environment)$ that is robust across all environments $\environmentSet$,
where $\inputModVec_\sample^\environment = \representFn(\inputOrgVec_\sample^\environment)$ is the vector of causal variables for the input $\inputOrgVec_\sample^\environment$ obtained through a representation function $\representFn(\cdot)$.
In this setting, we derive an invariant mapping that is simultaneously optimal over all environments \cite{Arjovsky2019}.
Formally, the design of the invariant predictor is cast as follows:
\begin{subequations}\label{eqn:IRM}
\begin{eqnarray}
\label{eqn:IRM_obj}
\underset{\representFn(\cdot),\modelParam}{\optMinimize} && \!\!\!\!\!\!\!\!
\textstyle
\sum\limits_{\environment\in\environmentSet} \frac{1}{\SAMPLE_\environment}
\sum\limits_{\sample}
\lossFn \big( \actionVecOpt^\environment_\sample, \mapping_{\modelParam}( \inputModVec^\environment_\sample ) \big), \\
\label{eqn:IRM_cns}
\subjectTo && \!\!\!\!\!\!\!\! 
\textstyle \modelParam \in \argmin_{\modelParam'} \frac{1}{\SAMPLE_\environment}
\sum_{\sample}
\lossFn \big( \actionVecOpt^\environment_\sample, \mapping_{\modelParam'}( \inputModVec^\environment_\sample ) \big) \,\, \forall \environment\in\environmentSet, \hphantom{tttt} \\
&& \!\!\!\!\!\!\!\!
\inputModVec^\environment_\sample = \representFn(\inputOrgVec^\environment_\sample) \qquad \forall (\inputOrgVec,\actionVecOpt)_\sample^\environment \in \dataSet.
\end{eqnarray}
\end{subequations}
It is worth highlighting that \emph{without \eqref{eqn:IRM_cns}}, the above problem is equivalent to \emph{multi-task learning} denoting environments as tasks, and solved using metalearning  or similar solutions \cite{Finn2017}.
However, with the simultaneous optimality introduced by \eqref{eqn:IRM_cns}, we seek for a different solution as discussed next.

Due to the constraints defined over environments, \eqref{eqn:IRM} boils down to a bilevel optimization problem that requires lower-level optimizations per environment, which makes solving \eqref{eqn:IRM} challenging.
Alternatively, following the \gls{irm} framework \cite{Arjovsky2019}, we recast \eqref{eqn:IRM_cns} as a penalized loss as follows: 
\begin{equation}\label{eqn:IRM_mod}
	\underset{\representFn(\cdot),\modelParam}{\optMinimize} \,
	\textstyle
	\sum\limits_{\environment\in\environmentSet} \frac{1}{\SAMPLE_\environment}
	\sum\limits_{\sample}
	\Big( 
	\underbrace{
	\lossFn \big( \actionVecOpt^\environment_\sample, \mapping_{\modelParam}( \inputModVec^\environment_\sample ) \big) 
	+ \regularizeCoef
	\| \grad{\modelParam}{ \lossFn \big( \actionVecOpt^\environment_\sample, \mapping_{\modelParam}( \inputModVec^\environment_\sample ) \big } \|_2
}_{F}
	\Big),
\end{equation}
where $\regularizeCoef>0$ is a constant hyperparameter.
By choosing a large $\regularizeCoef$, we enforce $	\| \grad{\modelParam}{ \lossFn \big( \actionVecOpt^\environment_\sample, \mapping_{\modelParam}( \inputModVec^\environment_\sample ) \big } \|_2 \to 0$ and thus, $	\grad{\modelParam}{ \lossFn \big( \actionVecOpt^\environment_\sample, \mapping_{\modelParam}( \inputModVec^\environment_\sample ) }  \approx 0$ constitutes the per-environment optimality defined under \eqref{eqn:IRM_cns}.

Based on the channel models \eqref{eqn:channel_ris_rx}-\eqref{eqn:channel_ris_tx_nlos}, it can be noticed that the composite \gls{csi} at the \gls{rx} depends on several features including distances, azimuth and elevation angles of \gls{tx}, \gls{rx}, and \gls{ris}.
Hence, the invariant properties of \glspl{aoa}, \glspl{aod}, and relative distances can be considered as an invariant representation such that $\inputModVec=\representFn(\inputOrgVec)$ of \gls{csi} irrespective of the environment in which the channels are generated.

Due to the nature of the choice of $\mapping_{\modelParam}(\cdot)$, determining the optimal $\modelParam$ could be derived analytically (e.g., $\mapping_{\modelParam}(\cdot)$ as a linear regression) or obtained via gradient decent methods (e.g., $\mapping_{\modelParam}(\cdot)$ as a \gls{nn}).
The objective of \eqref{eqn:IRM_mod} can be modified as follows:
\begin{multline}\label{eqn:modified_loss}
	F = 
	\lossCoef \textstyle
	\sum_{\element} 
	\sin^2 \big( \frac{ [\mapping_{\modelParam}(\inputModVec_\sample^\environment)]_\element - \actionOpt_{\element,\sample}^\environment}{2} ) + \\
	\textstyle
	\frac{\regularizeCoef\lossCoef}{2} \sqrt{ \sum_i
		\big( \sum_{\element} 
		\sin ( [\mapping_{\modelParam}(\inputModVec_\sample^\environment)]_\element - \actionOpt_{\element,\sample}^\environment ) \daba{[\mapping_{\modelParam}(\inputModVec_\sample^\environment)]_\element}{\modelParamInd_i} \big)^2 },
\end{multline}
by differentiating the loss function $\lossFn \big( \actionVecOpt^\environment_\sample, \mapping_{\modelParam}( \inputOrgVec^\environment_\sample ) \big)$ with respect to $\modelParam$.
Here, \eqref{eqn:modified_loss} is used as the loss function of the regression task of $\mapping_{\modelParam}(\cdot)$.
Then, using the training data $\dataSet$ gathered over multiple environments, we learn a \gls{nn}-based invariant phase predictor.

\begin{figure*}
\begin{minipage}{\linewidth}
	\centering
\subfloat[$\fractionEnv = 1$ and $\fractionCls=0.5$.]{
	\includegraphics[width=\myfigfactorx\linewidth]{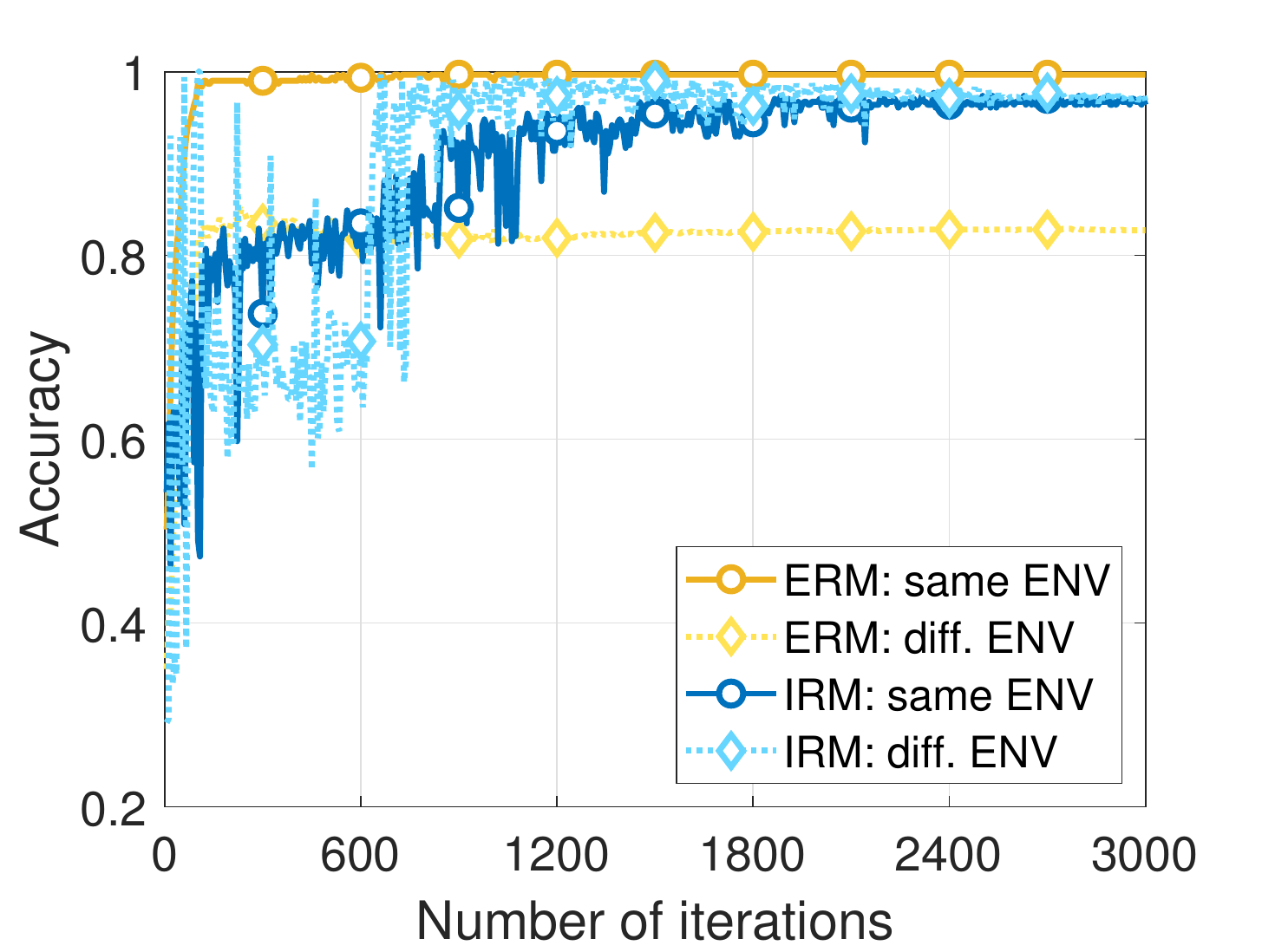}
	\label{fig:trainepochonenew}
}
\hspace{.005\linewidth}
\subfloat[$\fractionEnv = 0.5$ and $\fractionCls=0.5$.]{
	\includegraphics[width=\myfigfactorx\linewidth]{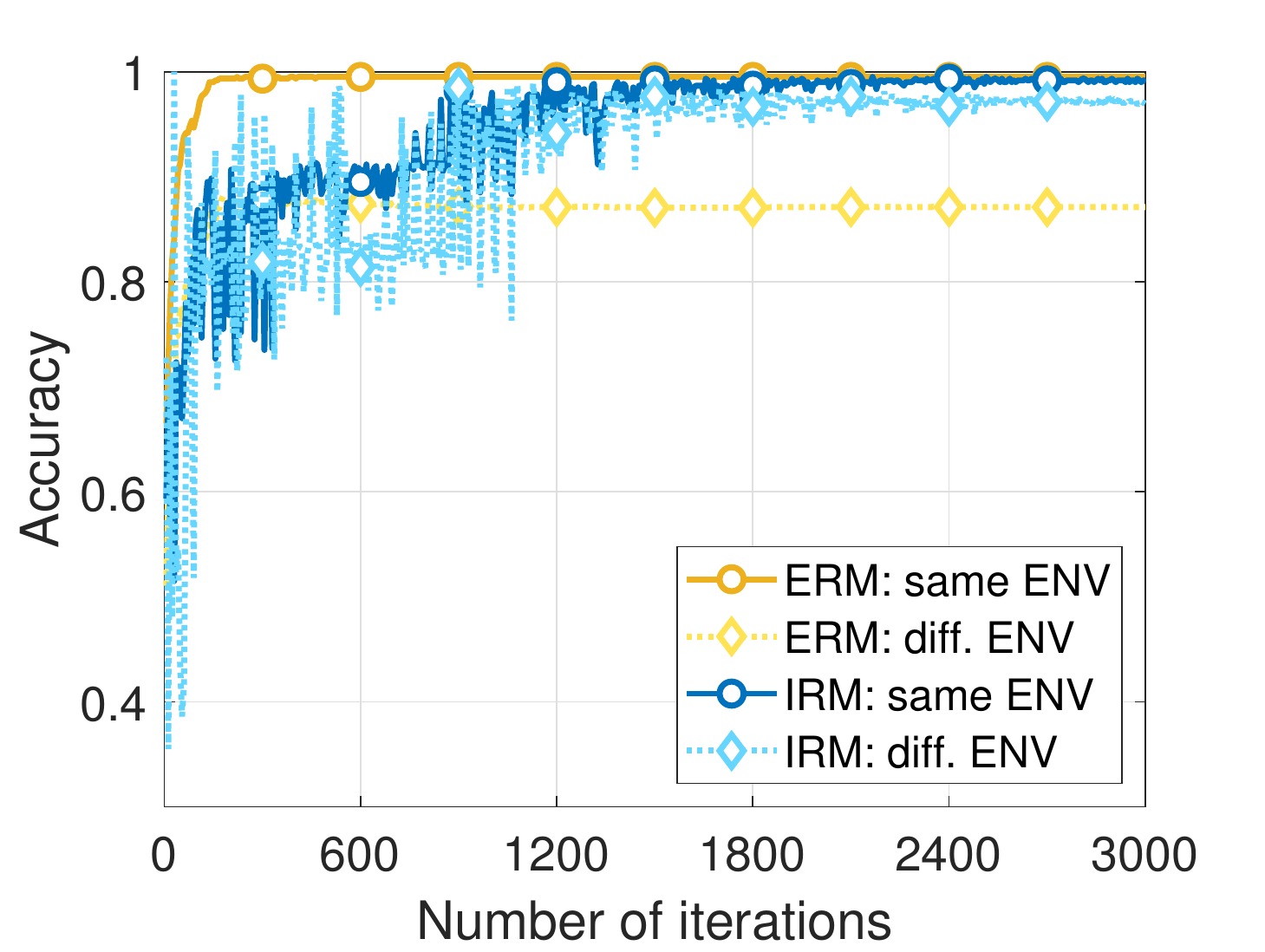}
	\label{fig:trainepochnew}
} 
\hspace{.005\linewidth}
\subfloat[$\fractionEnv = 0.5$ and $\fractionCls=0.8$.]{
	\includegraphics[width=\myfigfactorx\linewidth]{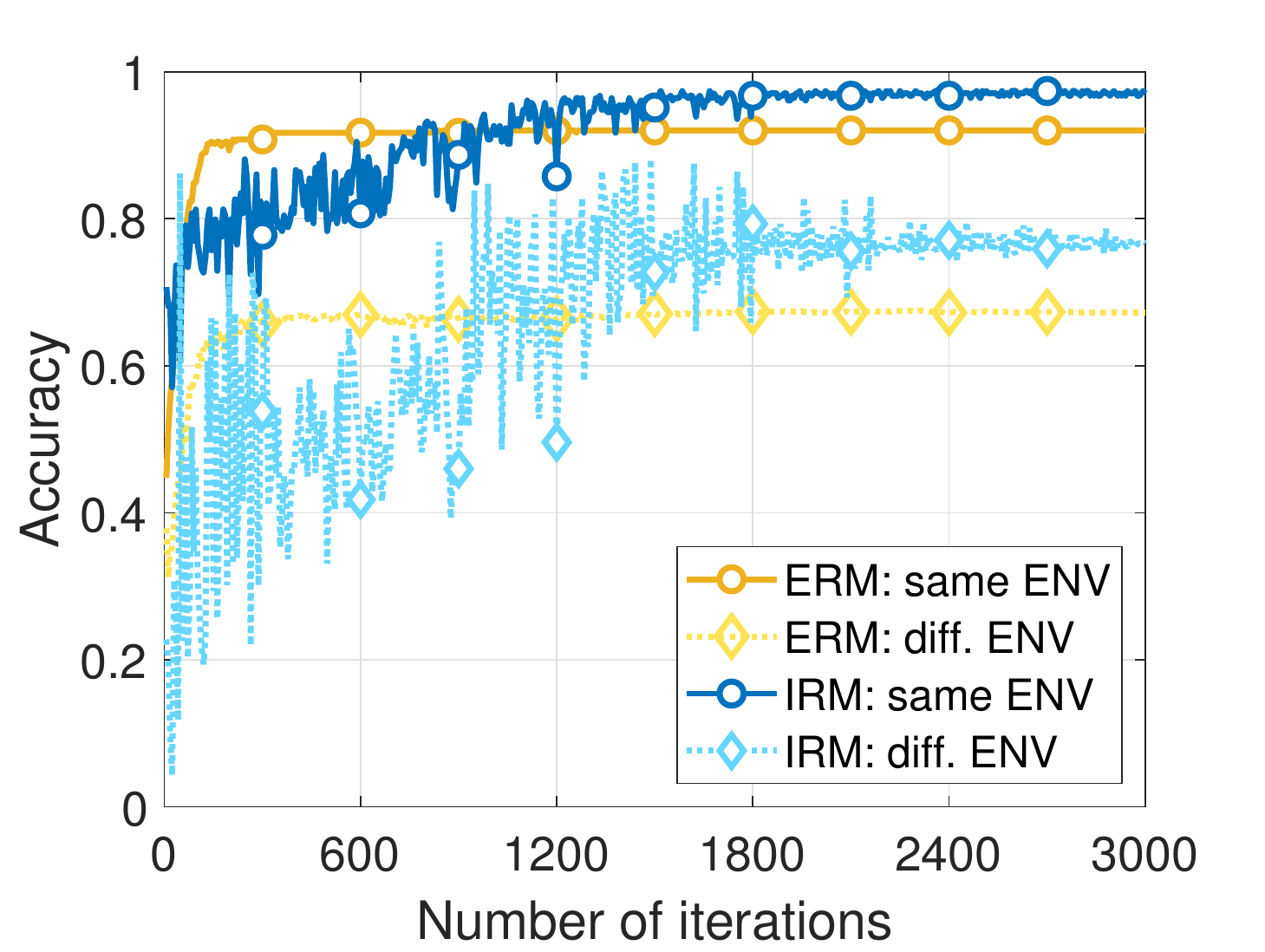}
	\label{fig:trainepochunbalancednew}
}
\caption{Training convergence of the proposed \irm{} and baseline \erm{} methods is compared within the \emph{same} environment (\env{1} and/or \env{2}) as training data as well as over a \emph{different} environment (\env{3}). }
\label{fig:trainepoch}
\end{minipage}

\vspace*{\floatsep}

\begin{minipage}{\linewidth}
	\centering
	\subfloat[Comparison of the prediction accuracy.]{
		\includegraphics[width=\myfigfactorx\linewidth]{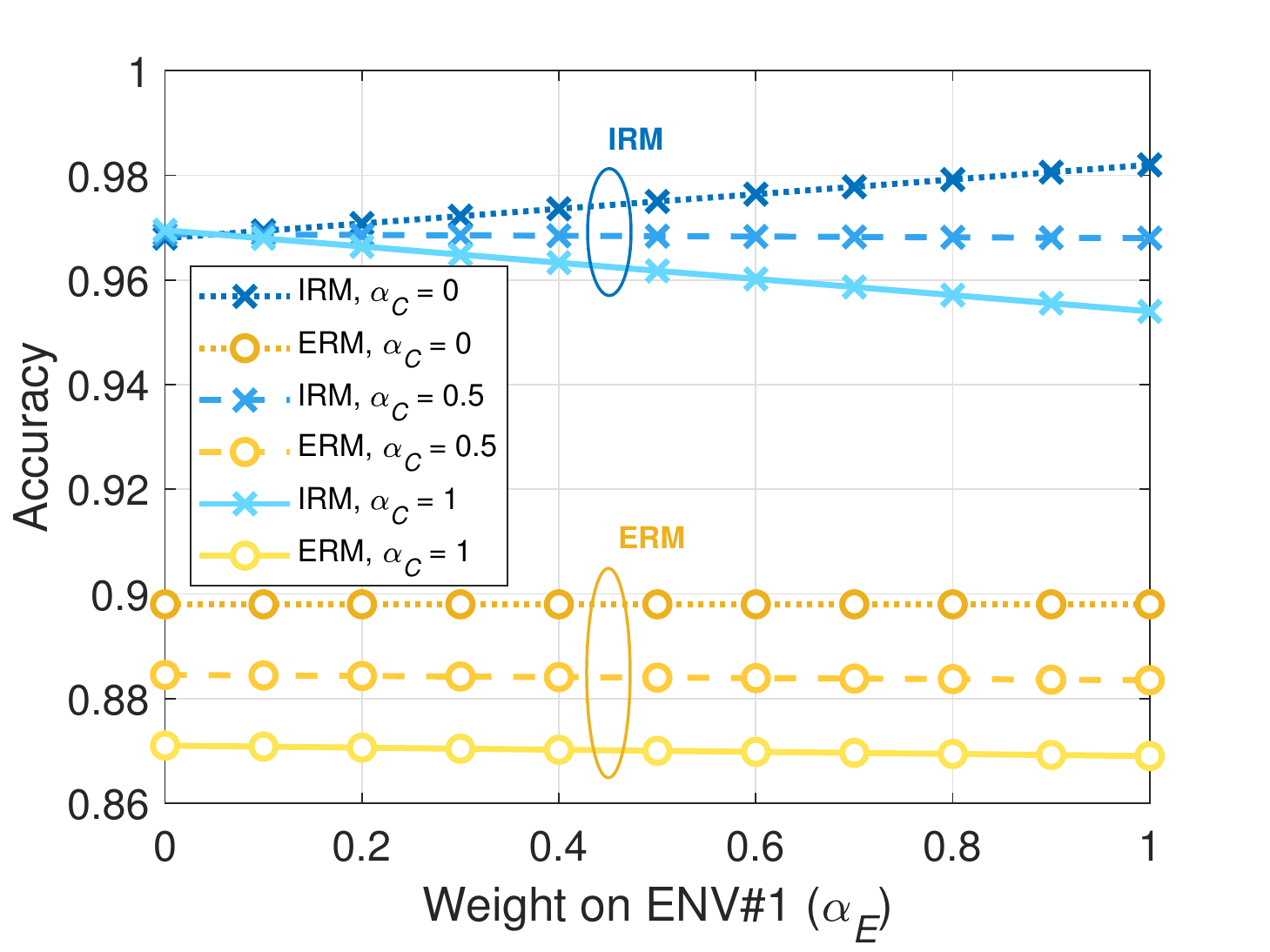}
		\label{fig:oodaccuracynew}
	}
	\hspace{.005\linewidth}
	\subfloat[Comparison of the \gls{snr} loss with $\fractionCls=0.5$.]{
		\includegraphics[width=\myfigfactorx\linewidth]{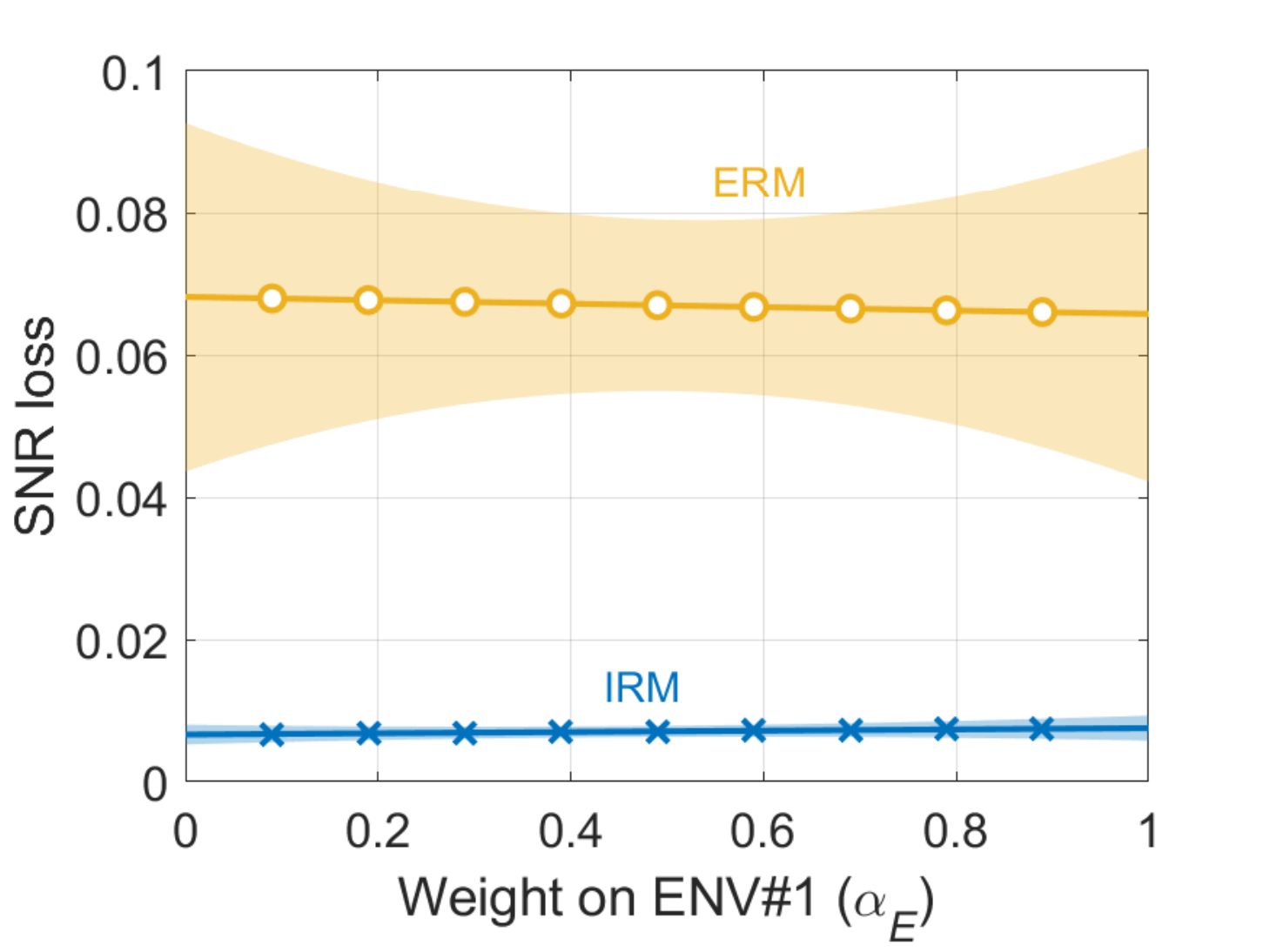}
		\label{fig:oodsnrenvnew}
	} 
	\hspace{.005\linewidth}
	\subfloat[Comparison of the \gls{snr} loss with $\fractionEnv=0.5$.]{
		\includegraphics[width=\myfigfactorx\linewidth]{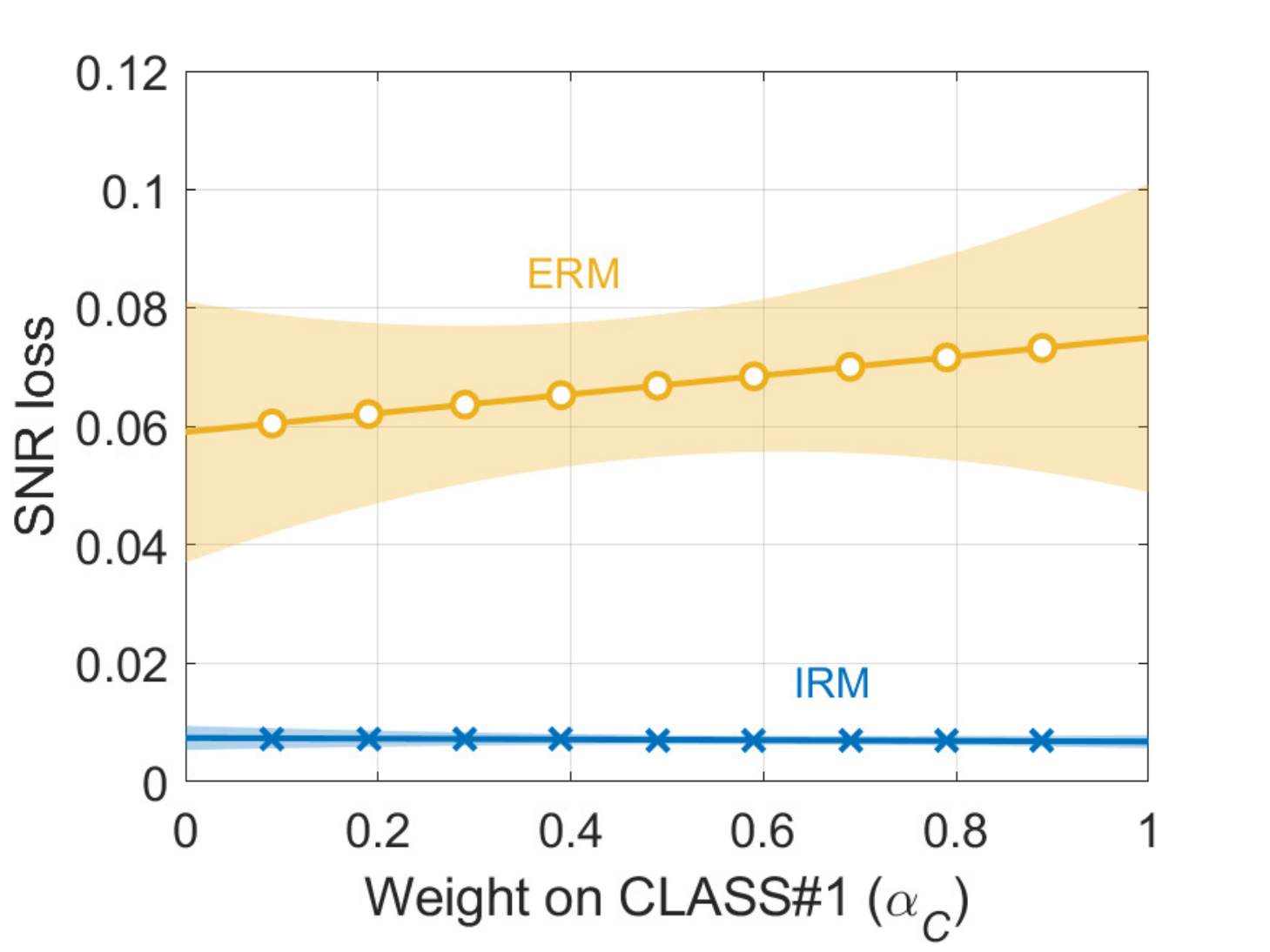}
		\label{fig:oodsnrclsnew}
	}
	\caption{Performance comparison of \erm{} and \irm{} methods for \gls{ood} testing datasets based on different choices of $\fractionEnv$ and $\fractionCls$. The shaded areas in right and middle figures correspond to the standard deviation of the \gls{snr} loss.}
	\label{fig:ood}
\end{minipage}
\end{figure*}

\section{Simulation Results}\label{sec:results}

The \gls{ris}-aided communication is simulated using the ``SimRIS'' channel simulator \cite{Basar2020}.
Therein, an indoor environment with the \gls{tx} located at the coordinate%
\footnote{
All distances are measured in meters within the Cartesian coordinates. 
}
$(0,35,1)$ using the 28\,GHz carrier frequency.
The \gls{ris} consists of $\ELEMENT = 10\times 10$ grid of reflective elements with half-wavelength inter-element distances located at $(10,30,1)$ facing the $y$-axis direction.
The \gls{rx} can be located at one of three environments, \env{1}, \env{2}, and \env{3}, such that the distances from \gls{ris} to the environment specific region $\mathcal{A}^{\environment}$ centers are 2\,m, 6\,m, and 4\,m, respectively.
From each environment, 1000 channel realizations are generated, \gls{csi} of incident and reflected signals decoupled over real and imaginary components ($\inputOrgVec\in\realset^{200\times 2}$) as well as \glspl{aod}, \glspl{aoa} (including azimuth and elevation), and distances of \gls{tx}-\gls{ris} and \gls{ris}-\gls{rx} ($\inputModVec\in\realset^{10}$) are recorded, and the configuration $\actionVecOpt$ yielding the highest \gls{snr} is labeled as CLASS\#1 if $\actionVecOpt=[0,0,\dots,0]$ and CLASS\#2 if $\actionVecOpt=[0,\pi,\dots,0,\pi]$.

For training, $\SAMPLE=600$ samples are collected from \env{1} and \env{2} while \env{3} is used as the testing dataset unless stated otherwise. 
The fraction of samples from \env{1} within the training dataset is denoted by $\fractionEnv$.
Out of two configuration classes, we select $\fractionCls:1-\fractionCls$ ratio of samples from CLASS\#1 and \#2 within \env{1} while $1-\fractionCls:\fractionCls$ ratio of samples from CLASS\#1 and \#2 within \env{2}.
Note that the testing dataset of \env{3} has equal number of samples from each class.
For both the baseline \erm{} as per \eqref{eqn:erm_problem} and the proposed method that solves \eqref{eqn:IRM_mod} (referred to as \irm{} hereinafter), \glspl{nn} with two fully-connected hidden layers of sizes $\{16,4\}$ within the \gls{mlp} architecture are trained as the phase predictors.
For the purpose of the benchmark, the ground truth results obtained via \emph{exhaustive search}, i.e., the best unique configurations $\actionVecOpt_\sample$ yielding the maximal rates, are referred to as \best{} while the random phase decision making is indicated by \random{}, hereinafter.
For sake of comparison, three different metrics are used for method $m\in\{\best, \irm, \erm, \random\}$:
(i) \emph{accuracy} = $\frac{1}{\SAMPLE}\sum_\sample \indictsimp{\actionVecEst_\sample=\actionVecOpt_\sample}$ measures the predictors' ability to infer the phase configuration as \best{} where $\indictsimp{\cdot}$ is the indicator function,
(ii) \emph{spectral efficiency} indicates the communication performance in terms of $\log_2(1 + \text{SNR}_{m})$,
and 
(iii) \emph{\gls{snr} loss} evaluates the communication degradation with respect to the \best{} case that is independent from the choices of $\txpower$, $\noiseAlone$, and $\bandwidth$, i.e., $\text{SNR loss} = 1 - \text{SNR}_{m} / \text{SNR}_{\text{\best}}$.

\textbf{Training Convergence and Inference Accuracy.}\quad
Within the same environment, \erm{} predictor achieves higher test accuracy (high generalization) over \irm{} when equal number of samples from all classes are used for the training as shown in Figs. \ref{fig:trainepochonenew} and \ref{fig:trainepochnew}. 
However, testing in a different environment (\env{3}) incurs a huge loss in accuracy under \erm{} highlights its lack of robustness.
In contrast, \irm{} yields slightly lower accuracy (low generalization) within the same environment under balanced datasets, yet
maintains high accuracy (high robustness) over \env{3} yielding beyond 95\% accuracy compared to 80-85\% with \erm{} as shown in Figs. \ref{fig:trainepochonenew} and \ref{fig:trainepochnew}.
When the dataset is biased towards specific classes, accuracy degradation (loss of generalization) is observed in both methods as illustrated in Fig. \ref{fig:trainepochunbalancednew}.
However, \irm{} predictor achieves about 78\% accuracy while \erm{} provides an accuracy of 67\%.

\textbf{\gls{ood} Robustness.}\quad
In Fig. \ref{fig:ood}, both \erm{} and \irm{} methods are trained over a dataset with $\fractionEnv=0.5$ and $\fractionCls=0.5$.
During testing, instead of using \env{3}, we used datasets \env{1} and \env{2} with different choices of $\fractionEnv$ and $\fractionCls$.
From Fig. \ref{fig:oodaccuracynew}, it can be noted that changing $\fractionEnv$ and $\fractionCls$ has a slight impact on \irm{} method for $\fractionCls=\{0,1\}$.
In contrast, the accuracy of \erm{} is consistent over $\fractionEnv$, yet much lower compared to \irm{} due to averaging of training samples.
Additionally, the \gls{snr} loss comparisons in Figs. \ref{fig:oodsnrenvnew} and \ref{fig:oodsnrclsnew} indicate that \irm{} not only results in about $90.2\%$ lower loss, but also yields about $86.1\%$ lower variance in errors compared to \erm{}. 
Compared to \irm{}, \erm{} exhibits higher error variance when the testing data deviates from the training data distribution.
The error variance when similar distributions are used for training and testing is about $1.2\times 10^{-2}$ while it increases to about $2.4\times 10^{-2}$ when the distribution is biased towards a single environment, and to about $2.5\times 10^{-2}$ under the bias towards a single class.
These comparisons further highlight the robustness of the proposed \irm{} design over the environment-unaware \erm{} method.

\begin{figure}
	\centering
	\includegraphics[width=\myfigfactor\linewidth]{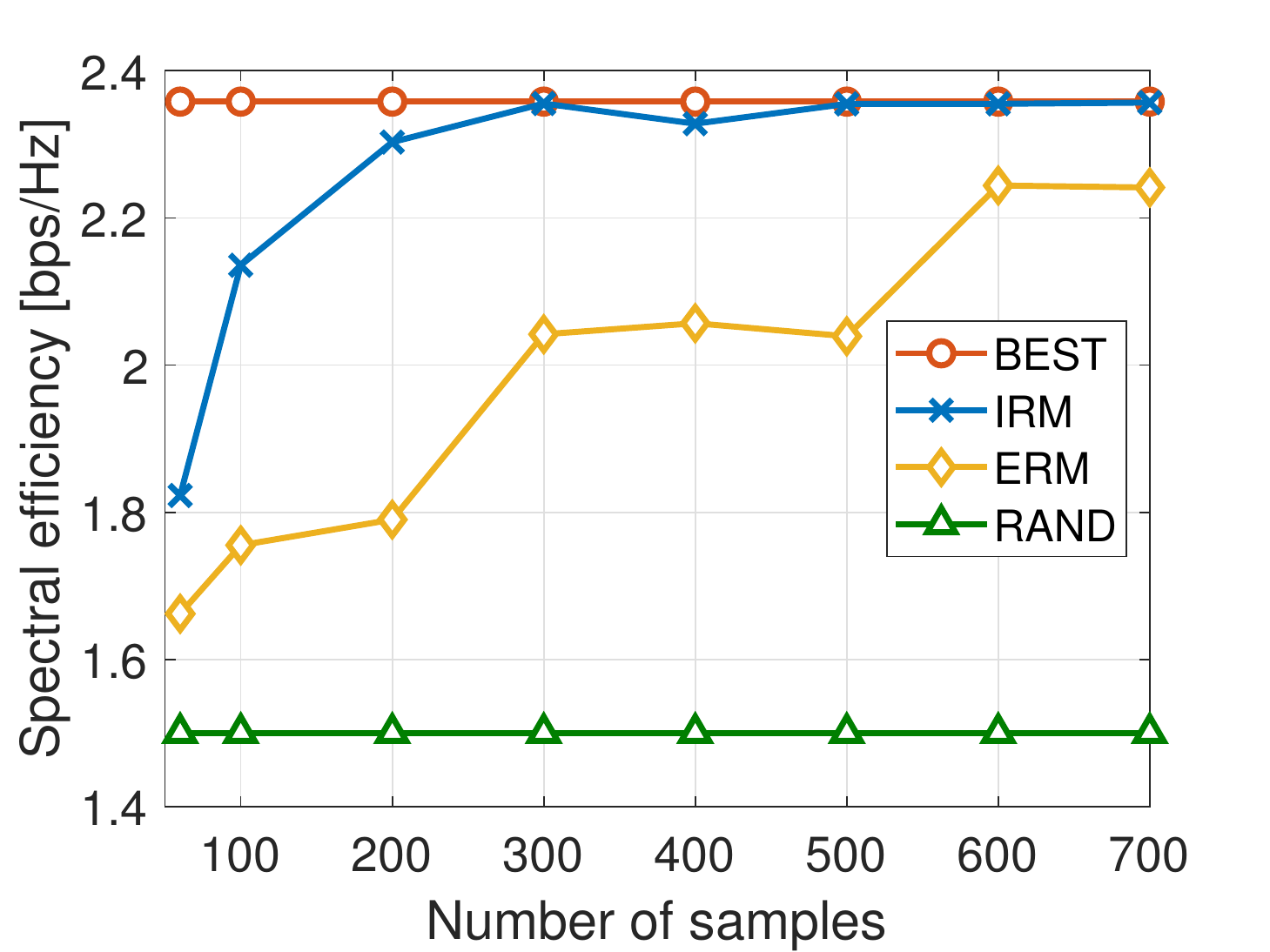}
	\caption{The impact of training sample complexity on the achievable spectral efficiency for all methods.}
	\label{fig:complexitynew}
	\vspace{-10pt}
\end{figure}

\textbf{Spectral Efficiency.}\quad
The achievable spectral efficiency in \env{3} with \best, \irm, \erm, and \random{} methods for different training sample complexities are compared in Fig. \ref{fig:complexitynew}.
Here, balanced training datsets are used, i.e., $\fractionEnv=0.5$ and $\fractionCls=0.5$. 
It can be noted that random choices result in 36.3\% loss of spectral efficiency compared to the best case.
With training over a small dataset consisting of 60 samples, \erm{} results in a loss of 29.5\% whereas for \irm{} the loss is 22.7\%.
As the dataset size is increased to 300 samples, the improvements in \erm{} results in a loss of 13.4\% whereas \irm{} yields a loss of 0.13\%.
Such close to the optimal performance with the proposed \irm{} predictor using fewer training samples compared to \erm{} is because of the learning over an invariant causal representation across multiple environments. 
Even with the increment of training samples to 700, \erm{} results in 4.3\% loss in terms of achievable spectral efficiency highlighting the lack of robustness using the \gls{erm} design.

\begin{figure}[!t]
	\centering
	\includegraphics[width=\myfigfactor\linewidth]{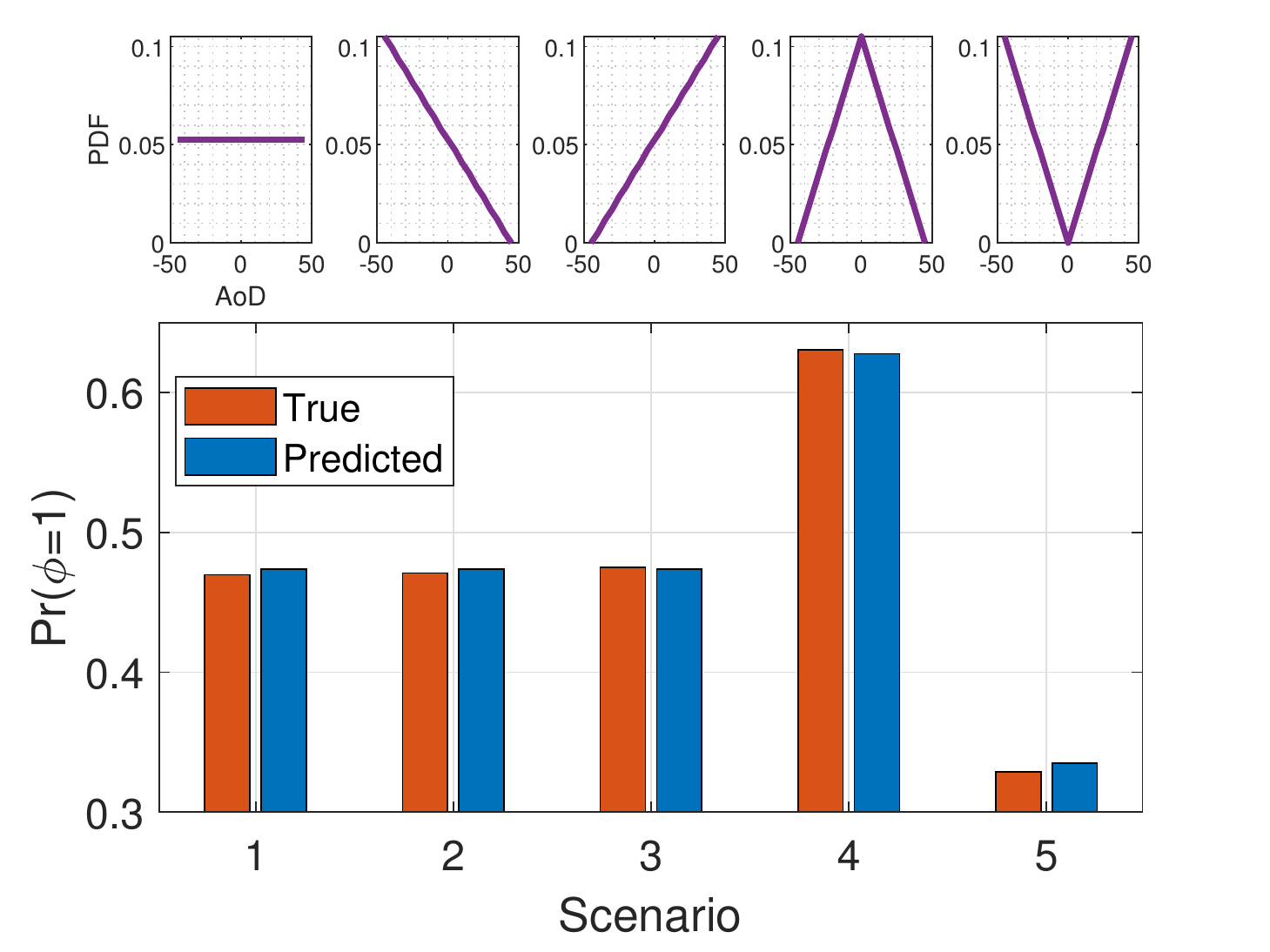}
	\caption{Validation of \irm's capability of predicting over interventions on \gls{aod} at the \gls{ris} following the distributions shown at the top.}
	\label{fig:interventions}
	\vspace{-10pt}
\end{figure}

\textbf{Impact of Causal Interventions.}\quad
Here, 
we define five different scenarios by intervening on the probabilistic generation process of the \gls{aod} at the \gls{ris} similar to Fig. \ref{fig:combined}c. 
In this view, we consider interventions that are constraining into 19 regular discrete steps spanning from $-50^\circ$ to $50^\circ$ following the distributions $\{\mathcal{P}_j\}_{j\in\{1,\dots,5\}}$ as illustrated at the top row of Fig. \ref{fig:interventions}.
Note that the \irm{} model is trained with a dataset consisting of \gls{aod} at the \gls{ris} (denoted by $z_1$ hereinafter) following a uniform distribution over $[-180^\circ,180^\circ]$, which we denote by $\mathcal{P}_0$. 
For each intervention $j$, we use \emph{do-calculus} \cite{Tucci2013} to compute the probability of the configuration being CLASS\#1, i.e., $\actionVecOpt = c_1$, by
\begin{equation}\label{eqn:doCalc}
	\probability(\actionVecEst=c_1) 
	=\textstyle  \sum_{z'} \mathcal{P}_0(\actionVecEst=c_1 | \text{do}(z_1=z')) \mathcal{P}_j(z'),
\end{equation}
where $\text{do}(z_1=z')$ represents the intervention taken with probability $\mathcal{P}_j(z')$ \textbf{rather than collecting data for the events} $z_1=z'$ with probability $\mathcal{P}_j(z')$.
For the numerical calculations, we generate $\SAMPLE'=\SAMPLE \mathcal{P}_j(z')$ samples and obtain the \gls{ris} configurations  using the \irm{} predictor, after which, $\mathcal{P}_0(\actionVecEst=c_1 | \text{do}(z_1=z')) = \frac{1}{\SAMPLE'}\sum_n \indictsimp{\actionVecEst=c_1}$ 
is computed.
The predicted value is then compared with the simulated outcome of \best{} method that uses a dataset generated based on $\mathcal{P}_j$ and the results are presented in Fig. \ref{fig:interventions} under the corresponding scenario $j$.
It can be noted that under the selected five scenarios, the predictions based on \irm{} coincide with the true probability of obtaining the CLASS\#1 under interventions.
The maximum deviation occurs with scenario 5 with a difference of $\sciNotation{6.3}{-3}$ in the probability underscoring the \irm's potential inference over interventions without the need for data collection or retraining.

\section{Conclusions}\label{sec:conclusion}

This paper investigates a novel system design for robust \gls{ris} based on leveraging the underlying causal structure that is invariant over different environments. 
The problem is cast as an \gls{irm} problem
as opposed to environment-unaware \gls{erm} approach.
Neural network-based phase configuration predictors are trained in a supervised learning manner and evaluated over different training and testing environments. 
The results indicate that the proposed \gls{irm}-based predictor is robust over \gls{erm} design across different environments and generalizes out-of-distribution.
Considering multiple transmitters, receivers, and \glspl{ris} with several antennas are interesting future extensions.  

\appendices

\bibliographystyle{IEEEtran}
\bibliography{IEEEabrv,mybib_ris_letter}

\begin{thebibliography}{10}
\providecommand{\url}[1]{#1}
\csname url@samestyle\endcsname
\providecommand{\newblock}{\relax}
\providecommand{\bibinfo}[2]{#2}
\providecommand{\BIBentrySTDinterwordspacing}{\spaceskip=0pt\relax}
\providecommand{\BIBentryALTinterwordstretchfactor}{4}
\providecommand{\BIBentryALTinterwordspacing}{\spaceskip=\fontdimen2\font plus
\BIBentryALTinterwordstretchfactor\fontdimen3\font minus
  \fontdimen4\font\relax}
\providecommand{\BIBforeignlanguage}[2]{{%
\expandafter\ifx\csname l@#1\endcsname\relax
\typeout{** WARNING: IEEEtran.bst: No hyphenation pattern has been}%
\typeout{** loaded for the language `#1'. Using the pattern for}%
\typeout{** the default language instead.}%
\else
\language=\csname l@#1\endcsname
\fi
#2}}
\providecommand{\BIBdecl}{\relax}
\BIBdecl

\bibitem{He2020}
J.~He, H.~Wymeersch, L.~Kong, O.~Silv{\'e}n, and M.~Juntti, ``Large intelligent
  surface for positioning in millimeter wave {MIMO} systems,'' in \emph{2020
  IEEE 91st Vehicular Technology Conference (VTC2020-Spring)}.\hskip 1em plus
  0.5em minus 0.4em\relax IEEE, 2020, pp. 1--5.

\bibitem{Oezdogan2020}
{\"O}.~{\"O}zdogan and E.~Bj{\"o}rnson, ``Deep learning-based phase
  reconfiguration for intelligent reflecting surfaces,'' \emph{preprint
  arXiv:2009.13988}, 2020.

\bibitem{Chen2019}
J.~Chen, Y.-C. Liang, H.~V. Cheng, and W.~Yu, ``Channel estimation for
  reconfigurable intelligent surface aided multi-user {MIMO} systems,''
  \emph{arXiv preprint arXiv:1912.03619}, 2019.

\bibitem{Gao2020}
J.~Gao, C.~Zhong, X.~Chen, H.~Lin, and Z.~Zhang, ``Unsupervised learning for
  passive beamforming,'' \emph{{IEEE} Commun. Lett.}, vol.~24, no.~5, pp.
  1052--1056, 2020.

\bibitem{Lee2020}
G.~Lee, M.~Jung, A.~T.~Z. Kasgari, W.~Saad, and M.~Bennis, ``Deep reinforcement
  learning for energy-efficient networking with reconfigurable intelligent
  surfaces,'' in \emph{Proc. of IEEE International Conference on Communications
  (ICC)}.\hskip 1em plus 0.5em minus 0.4em\relax IEEE, 2020, pp. 1--6.

\bibitem{Park2020}
J.~Park, S.~Samarakoon, H.~Shiri, M.~K. Abdel-Aziz, T.~Nishio, A.~Elgabli, and
  M.~Bennis, ``Extreme urllc: Vision, challenges, and key enablers,''
  \emph{arXiv preprint arXiv:2001.09683}, 2020.

\bibitem{Sheen2021}
B.~Sheen, J.~Yang, X.~Feng, and M.~M.~U. Chowdhury, ``A deep learning based
  modeling of reconfigurable intelligent surface assisted wireless
  communications for phase shift configuration,'' \emph{IEEE Open Journal of
  the Communications Society}, vol.~2, pp. 262--272, 2021.

\bibitem{Alexandropoulos2020}
G.~C. {Alexandropoulos} \emph{et~al.}, ``Phase configuration learning in
  wireless networks with multiple reconfigurable intelligent surfaces,'' in
  \emph{Proc. of 2020 IEEE Globecom Workshops}, 2020, pp. 1--6.

\bibitem{Basar2020}
E.~Basar, I.~Yildirim, and I.~F. Akyildiz, ``Indoor and outdoor physical
  channel modeling and efficient positioning for reconfigurable intelligent
  surfaces in {mmWave} bands,'' \emph{arXiv preprint arXiv:2006.02240}, 2020.

\bibitem{Arjovsky2019}
M.~Arjovsky, L.~Bottou, I.~Gulrajani, and D.~Lopez-Paz, ``Invariant risk
  minimization,'' \emph{arXiv preprint arXiv:1907.02893}, 2019.

\bibitem{Finn2017}
C.~Finn, P.~Abbeel, and S.~Levine, ``Model-agnostic meta-learning for fast
  adaptation of deep networks,'' in \emph{International Conference on Machine
  Learning}.\hskip 1em plus 0.5em minus 0.4em\relax PMLR, 2017, pp. 1126--1135.

\bibitem{Tucci2013}
R.~R. Tucci, ``Introduction to judea pearl's do-calculus,'' \emph{arXiv
  preprint arXiv:1305.5506}, 2013.

\end{thebibliography}

\end{document}